\documentclass{article}

\usepackage{lineno}
\modulolinenumbers[5]
\usepackage{rotating}

\usepackage{microtype}
\usepackage{graphicx}
\usepackage{subfigure}
\usepackage{booktabs} 

\usepackage{hyperref}
\usepackage{amsmath}
\DeclareMathOperator*{\argmin}{\arg\!\min}

\usepackage{xcolor,colortbl}
\usepackage{adjustbox}

\begin{document}

\title{A Comparative Analysis of XGBoost}



\author{Candice Bent\'ejac$^a$ \and Anna Cs\"org\H{o}$^b$ \and Gonzalo
Mart\'{\i}nez-Mu\~noz$^c$}
\date{%
    $^a$College of Science and Technology, University of Bordeaux, France\\%
    $^b$P\'azm\'any P\'eter Catholic University, Faculty of Information Technology and Bionics, Hungary\\
    $^b$Escuela Polit\'ectica Superior, Universidad Aut\'onoma de Madrid, Spain\\
}

\maketitle
\begin{abstract}
XGBoost is a scalable ensemble technique based on gradient boosting that has
demonstrated to be a reliable and efficient machine learning challenge solver.
This work proposes a practical analysis of how this novel technique works in
terms of training speed, generalization performance and parameter setup. In
addition, a comprehensive comparison between XGBoost, random forests and gradient
boosting has been performed using carefully tuned models as well as using
the default settings. The results of this comparison may indicate that XGBoost 
is not necessarily the best choice under all circumstances. Finally an extensive
analysis of XGBoost parametrization tuning process is carried out.
\end{abstract}

{\bf keywords: }
XGBoost, gradient boosting, random forest, ensembles of classifiers

\section{Introduction}
As machine learning is becoming a critical part of the success of more and more
applications --- such as credit scoring \cite{xia_2017_credit}, bioactive
molecule prediction \cite{babajide_2016_molecule}, solar and wind energy
prediction \cite{torres_2017_windsolar}, oil price prediction
\cite{gumus_2017_oil}, classification of galactic unidentified sources
\cite{mirabal_2016_galactic}, sentiment analysis \cite{valdivia18consensus}--- 
it is essential to find models that can deal efficiently with complex data, and with
large amounts of it. 
With that perspective in mind, ensemble methods have been a very effective tool
to improve the performance of multiple existing models
\cite{randomforests,gradientboosting,boosting,XGBoost}. These methods mainly
rely on randomization techniques, which consist in generating many diverse
solutions to the problem at hand 
\cite{randomforests}, or on adaptive emphasis procedures (e.g. boosting
\cite{boosting}). 

In fact, the above mentioned applications have in common that they all use ensemble
methods and, in particular, a recent
ensemble method called eXtreme Gradient Boosting or XGBoost \cite{XGBoost} with
very competitive results. This method, based on gradient boosting
\cite{gradientboosting}, has been consistently placing among the top contenders
in Kaggle competitions \cite{XGBoost}.
But XGBoost is not the only one to achieve remarkable results over a wide range
of problems. Random forest is also well known as one of the most accurate and as
a fast learning method independently from the nature of the datasets, as shown by
various recent comparative studies
\cite{classifiersarticle,caruana06empirical,rokach16df}.

This study follows the path of many other previous comparative analysis, such as
\cite{classifiersarticle,caruana06empirical,rokach16df}, with the intent of
covering a gap related to gradient boosting and its more recent variant XGBoost.
None of the previous comprehensive analysis included any machine learning
algorithm of the gradient boosting {\it family} despite of their appealing
properties.
The specific objectives of this study are, in the first place, to compare XGBoost's
performance with respect to the algorithm on which it is based (i.e. gradient
boosting). Secondly, the comparison is extended to random forest, which can be
considered as a benchmark since many previous comparisons demonstrated its
remarkable performance \cite{classifiersarticle,caruana06empirical}. 
The comparison is carried out in terms of accuracy and training speed. Finally,
a comprehensive analysis of the process of parameter setting in XGBoost is
performed. We believe this analysis can be very helpful to researchers of
various fields to be able to tune XGBoost more effectively.

The paper is organized as follows: Section 2 describes the methods of this
study, emphasizing the different parameters that need to be tuned; Section 3
presents the results of the comparison; Finally, the conclusions are summarized
in Section 4. 

\section{Methodology}
\subsection{Random forest}
One of the most successful machine learning methods is random forest \cite{randomforests}.
Random forest is an ensemble of classifiers composed of decision trees that
are generated using two different sources of randomization. First,
each individual decision tree is trained on a random sample with replacement
from the original data with the same size as the given training set. The
generated bootstrap samples are expected to have approximately $\approx 37\%$ of
duplicated instances. A second source of randomization applied in random forest
is attribute sampling. 
For that, at each node split, a subset of the input variables is randomly selected
to search for the best split. The value proposed by Breiman to be given to this
parameter is $\lfloor$log${}_2$(\#features)$+1\rfloor$.
For classification, the final prediction of the ensemble is given by majority
voting.

Based on the Strong Law of Large Numbers, it can be proven that the 
generalization error for random forests converges to a limit as the number of
trees in the forest becomes large \cite{randomforests}. The implication of this
demonstration is that the size of the ensemble is not a parameter that really
needs to be tuned, as the generalization accuracy of random forest does not
deteriorate on average when more classifiers are included into the ensemble.
The largest the
number of trees in the forest, the most probable the ensemble has converged to
its asymptotic generalization error.
Actually, one of the main advantages of random forest is that it is {\it almost}
parameter-free or at least, the default parameter setting has a remarkable
performance on average \cite{classifiersarticle}. The best two methods of 
that comparative study are based on random forest, for which only the value
of the number of random attributes that are selected at each split is tuned. The
method that placed fifth (out of 179 methods) in the comparison was random forest using
the default setting. This could also be seen as a drawback as it is difficult
to further improve random forest by parameter tuning.
 
Anyhow, other parameters that may be tuned in random forest are those
that control the depth of the decision trees. In general, decision trees in
random forest are grown until all leaves are pure. 
However, this can lead to very large trees. For such cases, the
growth of the tree can be limited by setting a maximum depth or by requiring a
minimum number of instances per node before or after the split.
 
Among the set of parameters that can be tuned for random forest, we evaluate the following 
ones in this study: 
\begin{itemize}
\item The number of features to consider when looking for the best split
(\verb?max_features?).
\item The minimum number of samples (\verb?min_samples_split?) required to split an
internal node. This parameter limits the size of the
trees but, in the worst case, the depth of the trees can be as large as
$N-\verb?min_samples_split?$, with $N$ the size of the training data.
\item The minimum number of samples (\verb?min_samples_leaf?) required
to create a leaf node. The effect of this limit is
different from the previous parameter, as it effectively removes split candidates
that are on the limits of the data distribution in the parent node. 
\item The maximum depth of the tree (\verb?max_depth?). This parameter limits
the depth of the tree independently of the number of instances that are in each
node.
\end{itemize}

\subsection{Gradient boosting}
Boosting algorithms combine weak learners, i.e. learners slightly better than 
random, into a strong learner in an iterative way \cite{boosting}.
Gradient boosting is a boosting-like algorithm for regression
\cite{gradientboosting}. 
Given a training dataset
$D=\{\mathbf{x}_i,y_i\}_{1}^N$, the goal of gradient boosting is to find an
approximation, $\hat{F}(\mathbf{x})$, of the function $F^*(\mathbf{x})$, which
maps instances $\mathbf{x}$ to their output values $y$, by minimizing the
expected value of a given loss function, $L(y, F(\mathbf{x}))$. 
Gradient boosting builds an additive approximation of $F^*(\mathbf{x})$ as 
a weighted sum of functions
\begin{equation}
F_m(\mathbf{x}) = F_{m-1}(\mathbf{x}) + \rho_m h_m(\mathbf{x}), 
\end{equation}
where $\rho_m$ is the weight of the $m^{th}$ function, $h_m(\mathbf{x})$. These
functions are the models of the ensemble (e.g. decision trees).
The approximation is constructed iteratively. First, a constant approximation of
$F^*(\mathbf{x})$ is obtained as
\begin{equation}
F_0(\mathbf{x}) = \argmin_{\alpha} \sum_{i=1}^N L(y_i, \alpha) \; .
\end{equation}
Subsequent models are expected to minimize 
\begin{equation}
(\rho_m, h_m(\mathbf{x})) = \argmin_{\rho,h} \sum_{i=1}^N L(y_i,
F_{m-1}(\mathbf{x}_i) + \rho h(\mathbf{x}_i))
\end{equation}
However, instead of solving the optimization problem directly, each $h_m$ can be
seen as a greedy step in a gradient descent optimization for $F^*$. For that,
each model, $h_m$, is trained on a new dataset
$D=\{\mathbf{x}_i,r_{mi}\}_{i=1}^N$, where the pseudo-residuals, $r_{mi}$, are
calculated by
\begin{equation}
r_{mi} = \left[ \frac{\partial L(y_i,F(\mathbf{x}))}{\partial F(\mathbf{x})}
\right]_{F(\mathbf{x})=F_{m-1}(\mathbf{x})}
\end{equation}
The value of $\rho_m$ is subsequently computed by solving a line search optimization
problem.

This algorithm can suffer from over-fitting if the iterative process is not
properly regularized \cite{gradientboosting}. For some loss
functions (e.g. quadratic loss), if the model $h_m$ fits the pseudo-residuals
perfectly, then in the
next iteration the pseudo-residuals become zero and the process terminates 
prematurely.
To control the additive process of gradient boosting, several regularization
parameters are considered. 
The natural way to regularize gradient boosting is to apply shrinkage to
reduce each gradient decent step
$F_m(\mathbf{x}) = F_{m-1}(\mathbf{x}) + \nu \rho_m h_m(\mathbf{x})$ with $\nu
=(0, 1.0]$. The value of $\nu$ is usually set to $0.1$.
In addition, further regularization can be achieved by limiting the complexity 
of the trained models. For the case of decision trees, we can limit the depth
of the trees or the minimum number of instances necessary to split a node.
Contrary to random forest, the default values for these parameters in gradient
boosting are set to harshly limit the expressive power of the trees (e.g. the depth is
generally limited to $\approx 3-5$).
Finally, another family of parameters also included in the different versions of
gradient boosting are those that randomize the base
learners, which can further improve the generalization of the ensemble
\cite{stochastic_gb}, such as random subsampling without replacement.

The attributes finally tested for gradient boosting are:
\begin{itemize}
\item The learning rate (\verb?learning_rate?) or shrinkage $\nu$.
\item The maximum depth of the three (\verb?max_depth?): the same meaning as in
the trees generated in random forest.
\item The subsampling rate (\verb?subsample?) for the size of the random
samples. Contrary to random forest, this is generally carried out without 
replacement \cite{stochastic_gb}. 
\item The number of features to consider when
looking for the best split (\verb?max_features?): as in random forest.
\item The minimum number of samples required to split an internal node
(\verb?min_samples_split?): as in random forest.
\end{itemize}

\subsection{XGBoost}
XGBoost \cite{XGBoost} is a decision tree ensemble based on gradient
boosting designed to be highly scalable.
Similarly to gradient boosting, XGBoost builds an additive expansion of the
objective function by minimizing a loss function. Considering that XGBoost is focused
only on decision trees as base classifiers, a variation of the loss function is
used to control the complexity of the trees
\begin{eqnarray}
L_{xgb} = \sum_{i=1}^N L(y_i, F(\mathbf{x}_i)) + \sum_{m=1}^M \Omega(h_m)\\
\Omega(h) = \gamma T + \frac{1}{2} \lambda \|w\|^2 \; ,
\end{eqnarray}
where $T$ is the number of leaves of the tree and $w$ are the output scores of
the leaves. This loss function can be integrated into the split criterion of
decision trees leading to a pre-pruning strategy. Higher values of $\gamma$ 
result in simpler trees. The value of $\gamma$ controls the
minimum loss reduction gain needed to split an internal node. An additional
regularization parameter in XGBoost is shrinkage, which reduces the step size
in the additive expansion. Finally, the complexity of the trees can also be
limited using other strategies as the depth of the trees, etc. A secondary
benefit of tree complexity reduction is that the 
models are trained faster and require less storage space.

Furthermore, randomization techniques are also implemented in XGBoost both to reduce
overfitting and to increment training speed. The randomization techniques
included in XGBoost are: random subsamples to train individual trees and column 
subsampling at tree and tree node levels.

In addition, XGBoost implements several methods to increment the training speed of
decision trees not directly related to ensemble accuracy. Specifically, XGBoost
focuses on reducing the computational
complexity for finding the best split, which is the most time-consuming part of
decision tree construction algorithms. Split finding algorithms usually
enumerate all possible candidate splits and select the one with the highest gain. 
This requires performing a linear scan over each sorted attribute to find the
best split for each node. To avoid sorting the data repeatedly in every node,
XGBoost uses a specific compressed column based structure in which the data is
stored pre-sorted. In this way, each attribute needs to be sorted only once.
This column based storing structure allows to find the best split for each
considered attributes in parallel. Furthermore, instead of scanning all possible
candidate splits, XGBoost implements a method based on percentiles of the data
where only a subset of candidate splits is tested and their gain is computed
using aggregated statistics. This idea resembles the node level data subsampling
that is already present in CART trees \cite{breiman84cart}. Moreover, a
sparsity-aware algorithm is used in XGBoost to effectively remove missing values
from the computation of the loss gain of split candidates.

The following parameters were tuned for XGBoost in this study:
\begin{itemize}
\item The learning rate (\verb?learning_rate?) or shrinkage $\nu$.
\item The minimum loss reduction (\verb?gamma?): The higher this value, the
shallower the trees. 
\item The maximum depth of the tree (\verb?max_depth?)
\item The fraction of features to be evaluated at each split
(\verb?colsample_bylevel?).
\item The subsampling rate (\verb?subsample?): sampling is done without
replacement. 
\end{itemize}

\section{Experimental results}

\begin{table}
\centering
\begin{tabular}{| c | c | c | c | c |}
  \hline
  \textbf{Name} 	& \textbf{Inst.} 	& \textbf{Attrs.} 	& \textbf{Miss.} 	&
\textbf{Class.}\\
  \hline
  Australia 	& 690 	& 14 	& Yes 	& 2 \\
  \hline
  Banknote 	& 1371 	& 5 	& No 	& 2 \\
  \hline
  Breast Cancer 	& 699 	& 10 	& Yes 	& 2 \\
  \hline
  Dermatology 	& 366 	& 33 	& Yes 	& 6 \\
  \hline
  Diabetes 	& 768 	& 20 	& Yes 	& 2 \\
  \hline
  Echo 	& 74 	& 12  	& Yes 	& 2 \\
  \hline
  Ecoli 	& 336 	& 8 	& No 	& 8 \\
  \hline
  German 	& 1000 	& 20 	& No 	& 2 \\
  \hline
  Heart 	& 270 	& 13 	& No 	& 2 \\
  \hline
  Heart Cleveland 	& 303 	& 75 	& Yes 	& 5 \\
  \hline
  Hepatitis 	& 155 	& 19 	& Yes 	& 2 \\
  \hline  
  Ionosphere 	& 351 	& 34 	& No 	& 2 \\
  \hline
  Iris 	& 150 	& 4 	& No 	& 3 \\
  \hline
  Liver 	& 583 	& 10 	& No 	& 2 \\
  \hline
  Magic04 	& 19020 	& 11 	& No 	& 2 \\
  \hline
  Parkinsons 	& 197 	& 23 	& No 	& 2 \\
  \hline
  Phishing 	& 1353 	& 10 	& No 	& 3 \\ 
  \hline
  Segment 	& 2310 	& 19 	& No 	& 7 \\ 
  \hline
  Sonar 	& 208 	& 60 	& No 	& 2 \\ 
  \hline
  Soybean 	& 675 	& 35 	& Yes 	& 18 \\
  \hline
  Spambase 	& 4601 	& 57 	& Yes 	& 2 \\
  \hline
  Teaching 	& 151 	& 5 	& No 	& 3 \\
  \hline
  Thyroid 	& 215 	& 21 	& No 	& 3 \\
  \hline
  Tic-Tac-Toe 	& 958 	& 9 	& No 	& 2 \\
  \hline  
  Vehicle 	& 946 	& 18 	& No 	& 4 \\
  \hline 
  Vowel 	& 990 	& 10 	& No 	& 11 \\
  \hline 
  Waveform 	& 5000 	& 21 	& No 	& 3 \\
  \hline
  Wine 	& 178 	& 13 	& No 	& 3 \\
  \hline
\end{tabular}
\caption{Characteristics of the studied datasets}
\label{table:datasets}
\end{table}

In this section, an extensive comparative analysis of the efficiency of random
forest, gradient boosting and XGBoost models is carried out. 
For the experiments, 28 different datasets coming from the UCI repository
\cite{UCI} were considered. These datasets come from different fields of
application, and have different number of attributes, classes and instances.
The characteristics of the analyzed datasets are shown in
Table~\ref{table:datasets}, which displays for each dataset its name,
number of instances, number of attributes, if the dataset has missing values and number of classes.
For this experiment, the implementation of \verb?scikit-learn? package
\cite{scikit-learn} was used for random forest and gradient boosting. For
XGBoost, the \verb?XGBoost? package
\footnote{\url{https://github.com/dmlc/xgboost}} was used.
In addition, for the comparison, XGBoost, random forest and gradient boosting
were analyzed tuning the parameters using a grid search as well as using the
default parameters of the corresponding packages.

The comparison was carried out using stratified 10-fold cross-validation.
For each dataset and partition of the data into train and test, the following
procedure was carried out for random forest, XGBoost and gradient boosting: 
(i) 
the optimum parameters for each method were estimated with stratified 10-fold
cross-validation within the training set using a grid search. A wide range of 
parameter values is explored in the grid search. For each of the three methods, 
these values are shown in Table~\ref{table:params}.
(ii) The best set of parameters extracted from the grid
search was used to train the corresponding ensemble using the whole training
partition; 
(iii) Additionally, for XGBoost, an ensemble was trained on the whole training
set for each possible combination of parameters given in Table~\ref{table:params}. This
allows us, in combination with step (i), to test for different grids (more details
down below);
(iv)
The default sets of parameters for each method were additionally used to train
an ensemble of each type (Table~\ref{table:params} shows the default values for
each parameter and ensemble type); 
(v)
The generalization accuracy of the three ensembles selected in (i) and of the
three ensembles with default parametrization is
estimated in the left out test set.
(v) In addition, the accuracy of all the XGBoost ensembles trained in step (iii)
is also computed using the test set.

All ensembles were composed of 200 decision trees. Note that the size of the
ensemble is not a parameter that needs to be tuned in ensembles of classifiers
\cite{gradientboosting,randomforests}. For random forest like ensembles, as more
trees are combined into the ensemble the generalization error tends to an
asymptotic value \cite{randomforests}. For gradient boosting like ensembles, the
generalization performance can deteriorate with the number of elements in the
ensemble especially for high learning rates values. However, this effect can not
only be be neutralized with the use of lower learning rates (or shrinkage) but
reverted \cite{gradientboosting}. The conclusion of \cite{gradientboosting} is
that the best option to regularize gradient boosting is to fix the number of
models to the highest computationally feasible value and to tune the learning
rate.
Although XGBoost has its own mechanism to handle missing values, we decided not
to use it in order to perform a fairer comparison with respect to random
forest and gradient boosting, 
as the implementation of decision trees in \verb?scikit-learn? does not handle
missing values. Instead, we used a class provided by \verb?scikit-learn? to
impute missing values by the mean of the whole attribute. 

\begin{table}
\centering
\begin{tabular}{| c | c | c |}
\hline
\multicolumn{3}{|c|}{\textbf{Random forest}}\\
\hline
\textbf{Parameter} 	& \textbf{Default value} 	& \textbf{Grid search values} \\
\hline
\verb?max_depth? 	& Unlimited 	& 5, 8, 10, unlimited\\
\hline
\verb?min_samples_split? 	& 2 	& 2, 5, 10, 20 \\
\hline
\verb?min_samples_leaf? 	& 1 	& 1, 25, 50,  70 \\
\hline
\verb?max_features? 	& sqrt 	& log2, 0.25,  sqrt, 1.0 \\
\hline
\multicolumn{3}{|c|}{\textbf{Gradient boosting}}\\
\hline
\textbf{Parameter} 	& \textbf{Default value} 	& \textbf{Grid search values} \\
\hline
\verb?learning_rate? 	& 0.1 	& 0.025, 0.05, 0.1, 0.2, 0.3 \\
\hline
\verb?max_depth? 	& 3 	& 2, 3, 5, 7, 10, unlimited \\
\hline
\verb?min_samples_split? 	& 2 	& 2, 5, 10, 20\\
\hline 
\verb?max_features? 	& 1.0 	& log2, sqrt, 0.25, 1.0 \\
\hline
\verb?subsample? 	& 1 	& 0.15, 0.5, 0.75, 1.0 \\
\hline
\multicolumn{3}{|c|}{\textbf{XGBoost}}\\
\hline
\textbf{Parameter} 	& \textbf{Default value} 	& \textbf{Grid search values} \\
\hline
\verb?learning_rate? 	& 0.1 	& 0.025, 0.05, 0.1, 0.2, 0.3 \\
\hline
\verb?gamma? 	& 0 	& 0, 0.1, 0.2, 0.3, 0.4, 1.0, 1.5, 2.0 \\
\hline
\verb?max_depth? 	& 3 	& 2, 3, 5, 7, 10, 100 \\
\hline
\verb?colsample_bylevel? 	& 1 	& log2, sqrt, 0.25, 1.0 \\ 
\hline
\verb?subsample? 	& 1 	& 0.15, 0.5, 0.75, 1.0 \\
\hline
\end{tabular}
\caption{Default values and possible values for every parameter in the
normal grid search for random forest, gradient
boosting and XGBoost} \label{table:params}
\end{table}

\subsection{Results}
Table~\ref{table:accuracy} displays the average accuracy and standard deviation
(after the $\pm$ sign) for: XGBoost with default parameters (shown with label D.
XGB), tuned XGBoost (as T. XGB in the table), random forest with default
parametrization (with D. RF), tuned random forest (T. RF), default gradient
boosting (D. GB) and tuned gradient boosting (T. GB). The best accuracy for each
dataset is highlighted using a yellow background. 

From Table~\ref{table:accuracy}, it can be observed that the method that obtains
the best performance in relation to the number of datasets with the highest
accuracy is tuned gradient
boosting (in 10 out of 28 datasets). After that, the methods in order are: tuned XGBoost,
that achieves the best results in 8 datasets; default gradient boosting in 5;
tuned and default random forest in 4; and default XGBoost in 3.
As it can be observed, the default performance of the three tested algorithms is
quite different. 
Default random forest is the method that performs more evenly with respect to
its tuned counterpart. Except for a few datasets, the differences in performance
are very small in random forest.
In fact, the difference between the tuned and the default parametrizations of random
forest is below $0.5\%$ in 18 out of 28 datasets.
Default XGBoost and gradient boosting perform generally worse than their tuned
versions. However, this is not always the case. 
This is especially evident in three cases for XGBoost ({\it German}, 
{\it Parkinson} and {\it Vehicle}), where the default parametrization for 
XGBoost achieves a better performance than the tuned XGBoost. 
These cases are a combination of two factors: noisy datasets and good default
settings. The parameter estimation process, even though it is performed within
train cross-validation, may overfit the training set specially in noisy datasets. 
In fact, it has been shown that reusing the training data multiple times can 
lead to overfitting \cite{dwork_2015_holdout}. On the other hand, in these
datasets, the default setting is one of the parametrizations that obtains the
best results both in train and test (among the best $\approx5\%$).

\begin{figure}
\centering
\rotatebox{270}{\includegraphics[scale=0.7]{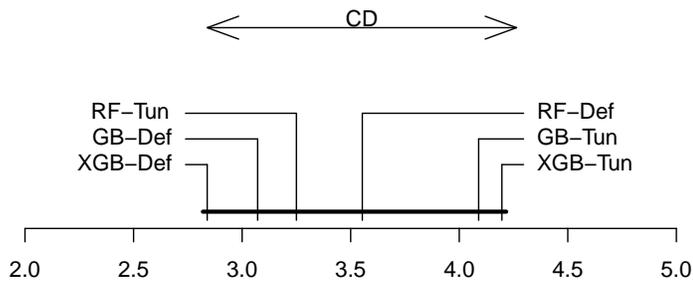}}
\caption{Average ranks (a higher rank is better) for the tested methods across 28
datasets (Critical difference CD= 1.42)}
\label{demsar}
\end{figure}

In order to summarize the figures shown in Table~\ref{table:accuracy},
we applied the methodology proposed in \cite{demsar2006}. 
This methodology compares the performance of several models across multiple
datasets. The comparison is carried out in terms of average rank of the
performance of each method in the tested datasets.
The results of this methodology are shown graphically in Figure~\ref{demsar}.
In this plot, a higher rank indicates a better accuracy. Statistical differences
among average ranks are determined using a Nemenyi test.
There are no statistical significant differences in average ranks between
methods connected with a horizontal solid line.
The critical distance over which 
the differences are considered significant is shown in the plot for reference
(CD = 1.42 for 6 methods, 28 datasets and p-value $<0.05$).

From Figure~\ref{demsar}, it can be observed that there are not any
statistically significant differences among the average ranks of the six tested
methods. The best method in
terms of average rank is tuned XGBoost, followed by gradient boosting and
random forest with default settings. 


\begin{sidewaystable}
\centering
\begin{adjustbox}{width=1\textwidth}
\begin{tabular}{| c | c | c | c | c | c | c |}
\hline
\textbf{Dataset} 	& \textbf{D. XGB} 	& \textbf{T. XGB} 	& \textbf{D. RF} 	& \textbf{T. RF} 	& \textbf{D. GB} 	& \textbf{T. GB}\\
\hline
Australian	& 86.94\%$\pm$2.73	& \cellcolor{yellow!47}87.53\%$\pm$3.47	& 87.26\%$\pm$3.83	& 86.08\%$\pm$3.42	& 86.23\%$\pm$3.41	& 86.38\%$\pm$3.39\\
\hline
Banknote	& 99.64\%$\pm$0.59	& 99.64\%$\pm$0.49	& 99.34\%$\pm$0.60	& 99.13\%$\pm$0.71	& 99.71\%$\pm$0.48	& \cellcolor{yellow!47}99.78\%$\pm$0.33\\
\hline
Breast	& 96.28\%$\pm$1.15	& 95.99\%$\pm$1.68	& \cellcolor{yellow!47}96.99\%$\pm$1.36	& 96.85\%$\pm$1.26	& 96.71\%$\pm$1.58	& 96.42\%$\pm$1.61\\
\hline
Cleveland	& 81.14\%$\pm$7.25	& 83.16\%$\pm$7.94	& 82.46\%$\pm$6.77	& 82.46\%$\pm$8.03	& \cellcolor{yellow!47}83.78\%$\pm$6.48	& 82.16\%$\pm$8.35\\
\hline
Dermatology	& 96.74\%$\pm$3.21	& 97.27\%$\pm$3.24	& 97.27\%$\pm$3.29	& \cellcolor{yellow!47}97.30\%$\pm$3.24	& 96.47\%$\pm$3.28	& 96.18\%$\pm$2.81\\
\hline
Diabetes	& 75.65\%$\pm$5.11	& 76.56\%$\pm$4.50	& 76.69\%$\pm$3.45	& 76.69\%$\pm$4.89	& \cellcolor{yellow!47}76.81\%$\pm$4.43	& 76.30\%$\pm$3.74\\
\hline
Echo	& 94.46\%$\pm$6.80	& \cellcolor{yellow!47}98.75\%$\pm$3.75	& \cellcolor{yellow!47}98.75\%$\pm$3.75	& 97.32\%$\pm$5.37	& 97.32\%$\pm$5.37	& 95.89\%$\pm$6.29\\
\hline
Ecoli	& 86.87\%$\pm$5.31	& 89.05\%$\pm$4.12	& 89.07\%$\pm$5.00	& \cellcolor{yellow!47}89.11\%$\pm$4.71	& 87.25\%$\pm$6.88	& 87.81\%$\pm$4.77\\
\hline
German	& \cellcolor{yellow!47}79.00\%$\pm$4.22	& 77.40\%$\pm$4.13	& 76.40\%$\pm$4.48	& 75.80\%$\pm$4.17	& 76.70\%$\pm$5.12	& 77.20\%$\pm$3.99\\
\hline
Heart	& 79.26\%$\pm$5.29	& 84.07\%$\pm$5.98	& 83.33\%$\pm$5.56	& \cellcolor{yellow!47}84.44\%$\pm$5.19	& 81.85\%$\pm$7.67	& 83.70\%$\pm$5.29\\
\hline
Hepatitis	& 59.21\%$\pm$8.28	& \cellcolor{yellow!47}67.00\%$\pm$6.56	& 65.54\%$\pm$12.35	& 61.83\%$\pm$12.68	& 56.58\%$\pm$11.90	& 64.96\%$\pm$13.08\\
\hline
Ionosphere	& 92.56\%$\pm$2.69	& 92.59\%$\pm$3.17	& 93.44\%$\pm$2.88	& 93.16\%$\pm$2.91	& \cellcolor{yellow!47}93.72\%$\pm$2.51	& 92.85\%$\pm$3.02\\
\hline
Iris	& 92.67\%$\pm$6.29	& 94.00\%$\pm$4.67	& \cellcolor{yellow!47}94.67\%$\pm$4.99	& 92.67\%$\pm$6.29	& \cellcolor{yellow!47}94.67\%$\pm$4.99	& \cellcolor{yellow!47}94.67\%$\pm$4.99\\
\hline
Liver	& 68.65\%$\pm$4.69	& 68.11\%$\pm$6.12	& 67.76\%$\pm$5.03	& 67.58\%$\pm$4.07	& 69.33\%$\pm$5.26	& \cellcolor{yellow!47}70.16\%$\pm$4.42\\
\hline
Magic04	& 87.47\%$\pm$0.57	& 88.63\%$\pm$0.48	& 88.19\%$\pm$0.42	& 88.18\%$\pm$0.46	& 86.85\%$\pm$0.37	& \cellcolor{yellow!47}88.83\%$\pm$0.45\\
\hline
New-thyroid	& 95.80\%$\pm$3.26	& 95.80\%$\pm$3.26	& \cellcolor{yellow!47}96.75\%$\pm$2.94	& 96.28\%$\pm$3.48	& \cellcolor{yellow!47}96.75\%$\pm$2.94	& 96.28\%$\pm$3.48\\
\hline
Parkinsons	& \cellcolor{yellow!47}92.14\%$\pm$5.72	& 90.14\%$\pm$4.01	& 90.70\%$\pm$5.60	& 90.70\%$\pm$4.62	& 91.14\%$\pm$6.00	& 91.20\%$\pm$4.23\\
\hline
Phishing	& 89.65\%$\pm$2.44	& \cellcolor{yellow!47}91.13\%$\pm$1.30	& 88.63\%$\pm$3.38	& 89.51\%$\pm$2.42	& 90.62\%$\pm$1.76	& 90.25\%$\pm$2.19\\
\hline
Segment	& 98.48\%$\pm$0.70	& \cellcolor{yellow!47}98.70\%$\pm$0.77	& 98.01\%$\pm$0.80	& 98.18\%$\pm$0.67	& 97.75\%$\pm$0.74	& 98.35\%$\pm$0.64\\
\hline
Sonar	& 85.59\%$\pm$6.44	& 86.97\%$\pm$4.84	& 83.64\%$\pm$3.87	& 85.59\%$\pm$4.83	& 84.66\%$\pm$4.61	& \cellcolor{yellow!47}88.95\%$\pm$4.32\\
\hline
Soybean	& 94.80\%$\pm$3.35	& \cellcolor{yellow!47}95.22\%$\pm$2.96	& 94.06\%$\pm$2.27	& 94.65\%$\pm$2.57	& 93.63\%$\pm$2.86	& 93.58\%$\pm$3.45\\
\hline
Spambase	& 95.17\%$\pm$1.29	& 95.57\%$\pm$1.28	& 95.46\%$\pm$1.38	& 95.48\%$\pm$1.26	& 94.59\%$\pm$1.43	& \cellcolor{yellow!47}96.11\%$\pm$1.20\\
\hline
Teaching	& 63.55\%$\pm$8.00	& 64.26\%$\pm$14.15	& 65.51\%$\pm$8.95	& 63.41\%$\pm$11.13	& 62.76\%$\pm$7.38	& \cellcolor{yellow!47}68.75\%$\pm$14.65\\
\hline
Tic-tac-toe	& 96.56\%$\pm$1.63	& \cellcolor{yellow!47}100.00\%$\pm$0.00	& 95.52\%$\pm$1.68	& 95.62\%$\pm$1.38	& 90.09\%$\pm$2.77	& \cellcolor{yellow!47}100.00\%$\pm$0.00\\
\hline
Vehicle	& \cellcolor{yellow!47}78.74\%$\pm$3.29	& 77.18\%$\pm$2.31	& 74.50\%$\pm$3.55	& 74.13\%$\pm$3.98	& 78.15\%$\pm$1.89	& 77.79\%$\pm$2.04\\
\hline
Vowel	& 92.63\%$\pm$3.53	& 95.86\%$\pm$2.14	& 97.47\%$\pm$1.37	& \cellcolor{yellow!47}97.78\%$\pm$1.68	& 93.23\%$\pm$3.32	& 96.77\%$\pm$2.01\\
\hline
Waveform	& 85.72\%$\pm$1.05	& 85.72\%$\pm$1.77	& 85.42\%$\pm$1.73	& 85.62\%$\pm$1.46	& 85.32\%$\pm$0.85	& \cellcolor{yellow!47}85.86\%$\pm$1.18\\
\hline
Wine	& 97.18\%$\pm$2.83	& \cellcolor{yellow!47}98.82\%$\pm$2.37	& 98.26\%$\pm$2.66	& 98.26\%$\pm$2.66	& 97.74\%$\pm$2.78	& \cellcolor{yellow!47}98.82\%$\pm$2.37\\
\hline
\end{tabular}
\end{adjustbox}
\caption{Average accuracy and standard deviation for random forest, 
gradient boosting and XGBoost, both using default and tuned parameter settings
} \label{table:accuracy}
\end{sidewaystable}

In Table~\ref{table:time}, the average training execution time (in seconds) for
the analyzed datasets is shown. For the methods using the default settings, the
table shows the training time. For the methods that have been tuned using
grid search, two numbers are shown separated with a `+' sign. The first number
corresponds to the time spent in the within-train 10-fold cross-validated grid
search.
The second figure corresponds to the
training time of each method, once the optimum parameters have been estimated.
The last row of the table reports the average ratio of each execution
time with respect to the execution time of XGBoost using the default setting.
All experiments were run using an eight-core          
Intel$^{\mbox{\scriptsize{\textregistered}}}$ Core                               
$^{\mbox{\scriptsize{\texttrademark}}}$ i7-4790K CPU @ 4.00GHz processor.
The reported times are sequential times in order to have a real measure of the
needed computational resources, even though the grid search was performed in 
parallel using all available cores. This comparison is fair independently of
whether the
learning algorithms include internal multi-thread optimizations or not.
For instance random forest and XGBoost include multi-thread optimizations in their code 
to compute the splits in XGBoost and to train each single tree in random forest
whereas gradient boosting does not. 
Notwithstanding, given that the grid
search procedure is fully parallelizable, these optimizations do not reduce 
the end-to-end time required to perform a grid search in a real setting.

\begin{table}
\centering
\begin{adjustbox}{width=1\textwidth}
\begin{tabular}{| c | c | c | c | c | c | c |}
\hline
\textbf{Dataset} 	& \textbf{D. XGB} 	& \textbf{T. XGB} 	& \textbf{D. RF} 	&
\textbf{T. RF} 	& \textbf{D. GB} 	& \textbf{T. GB} \\
\hline
Australian 	& 0.10 	& 2775 + 0.08 	& 0.32 	& 766 + 0.30 	& 0.12 	& 6053 + 0.61\\
\hline
Banknote 	& 0.10 	& 2874 + 0.06 	& 0.40 	& 991 + 0.39 	& 0.23 	& 5410 + 0.15\\
\hline
Breast 	& 0.06 	& 1733 + 0.04 	& 0.28 	& 744 + 0.28 	& 0.16 	& 4071 + 0.35\\
\hline
Cleveland 	& 0.05 	& 1357 + 0.02 	& 0.28 	& 700 + 0.26 	& 0.10 	& 3676 + 0.10\\
\hline
Dermatology 	& 0.46 	& 11014 + 0.27 	& 0.27 	& 727 + 0.27 	& 0.76 	& 17840 + 0.93\\
\hline
Diabetes 	& 0.08 	& 2982 + 0.06 	& 0.35 	& 798 + 0.34 	& 0.13 	& 6612 + 0.50\\
\hline
Echo 	& 0.02 	& 532 + 0.01 	& 0.25 	& 671 + 0.25 	& 0.05 	& 1608 + 0.08\\
\hline
Ecoli 	& 0.17 	& 4518 + 0.11 	& 0.27 	& 715 + 0.28 	& 0.67 	& 15448 + 1.25\\
\hline
German 	& 0.17 	& 5453 + 0.16 	& 0.37 	& 790 + 0.37 	& 0.13 	& 8589 + 0.48\\
\hline
Heart 	& 0.05 	& 1268 + 0.03 	& 0.27 	& 716 + 0.27 	& 0.12 	& 3579 + 0.10\\
\hline
Hepatitis 	& 0.04 	& 1071 + 0.02 	& 0.26 	& 686 + 0.26 	& 0.11 	& 2756 + 0.11\\
\hline
Ionosphere 	& 0.14 	& 2487 + 0.05 	& 0.34 	& 893 + 0.35 	& 0.27 	& 3982 + 0.22\\
\hline
Iris 	& 0.04 	& 1308 + 0.03 	& 0.25 	& 678 + 0.25 	& 0.29 	& 6555 + 0.39\\
\hline
Liver 	& 0.07 	& 2308 + 0.05 	& 0.35 	& 804 + 0.31 	& 0.16 	& 5559 + 0.28\\
\hline
Magic04 	& 3.32 	& 123764 + 7.86 	& 11.39 	& 11860 + 9.44 	& 1.50 	& 160831 + 48.07\\
\hline
New-thyroid 	& 0.05 	& 1728 + 0.04 	& 0.26 	& 686 + 0.26 	& 0.30 	& 7426 + 0.42\\
\hline
Parkinsons 	& 0.05 	& 1169 + 0.03 	& 0.27 	& 697 + 0.28 	& 0.10 	& 2539 + 0.11\\
\hline
Phishing 	& 0.36 	& 12464 + 0.75 	& 0.33 	& 816 + 0.37 	& 0.80 	& 31504 + 1.07\\
\hline
Segment 	& 3.22 	& 73309 + 2.21 	& 0.77 	& 1451 + 0.79 	& 2.25 	& 62647 + 2.71\\
\hline
Sonar 	& 0.15 	& 2442 + 0.05 	& 0.31 	& 828 + 0.33 	& 0.32 	& 3729 + 0.28\\
\hline
Soybean 	& 3.73 	& 104873 + 2.69 	& 0.31 	& 762 + 0.31 	& 3.14 	& 75660 + 5.51\\
\hline
Spambase 	& 1.83 	& 47412 + 1.46 	& 1.40 	& 2016 + 0.73 	& 0.32 	& 33088 + 4.79\\
\hline
Teaching 	& 0.05 	& 1879 + 0.06 	& 0.26 	& 676 + 0.26 	& 0.35 	& 8955 + 0.35\\
\hline
Tic-tac-toe 	& 0.08 	& 2842 + 0.08 	& 0.32 	& 746 + 0.32 	& 0.13 	& 8184 + 0.45\\
\hline
Vehicle 	& 0.60 	& 17217 + 0.38 	& 0.43 	& 843 + 0.42 	& 0.70 	& 24987 + 0.88\\
\hline
Vowel 	& 1.96 	& 48726 + 1.24 	& 0.57 	& 977 + 0.50 	& 2.31 	& 61085 + 7.67\\
\hline
Waveform 	& 3.34 	& 111562 + 1.74 	& 2.46 	& 3268 + 1.58 	& 1.66 	& 96197 + 11.44\\
\hline
Wine 	& 0.07 	& 2104 + 0.05 	& 0.26 	& 700 + 0.26 	& 0.28 	& 7197 + 0.33\\
\hline
\hline
Ave. ratio 	& 1.0 	& 29043.7 + 0.8 	& 3.6 	& 8953.4 + 3.5 	& 2.4 	& 69238.2 + 4.3 \\ 
\hline
\end{tabular}
\end{adjustbox}
\caption{Average execution time (in seconds) for training XGBoost,
random forest and gradient boosting (more details in the text)} \label{table:time}
\end{table}

As it can be observed from Table~\ref{table:time}, finding the best parameters
to tune the classifiers through the grid search is a rather costly process. 
In fact, the end-to-end training time of the tuned models is clearly dominated by the 
grid search process, which contributes with a percentage over 99.9\% to the training
time. 
Since the size of the grid is different for different classifiers (i.e. 3840,
256 and 1920 for XGB, RF and GB respectively), the time dedicated to finding 
the best parameters is not directly comparable between classifiers.

However, when it comes to fitting a single ensemble to the training data
without taking into account the grid search time, XGBoost clearly shows the 
fastest performance on average. The time necessary to train XGBoost given a set
of parameters is about 3.5 times faster than training a random forest and 2.4-4.3
times faster than training a gradient boosting model. This last difference can 
be observed in the time employed in the grid search by XGBoost and gradient
boosting. XGBoost takes less than half of the time to look for the best
parameter setting than gradient boosting, despite the fact that its grid size
is twice the size of the grid of gradient boosting.
Finally, for some multiclass problems, as {\it Segment} or
{\it Soybean}, the execution time of XGBoost and gradient boosting deteriorates
in relation to random forest. 

\subsection{Analysis of XGBoost parametrization}
In order to further analyze and understand the parametrization of XGBoost, we
carry out further experiments with two objectives. The first objective is
to try to select a better default parametrization for XGBoost. The second is to
explore alternative grids for XGBoost. For the first objective, we have analyzed,
 for each parameter, the relation among single value assignments. To do so, we have computed
the average rank (in test error) across all datasets for the ensembles trained
using all parameter configurations as given in Table~\ref{table:params} for
XGBoost. Recall that we have analyzed 3840 possible parameter configurations.
Then, for each given parameter, say {\it ParamX}, and parameter value, say {\it
ParamX\_valueA}, we compute the percentage of times that the average rank
improves when {\it ParamX=ParamX\_valueA} is changed to another value 
{\it ParamX\_valueB} and no other parameter is modified. These results are
shown in Tables~\ref{table:l} to \ref{table:ss}. The tables are to be read as follows:
each cell of the table indicates the \% of times the average rank 
improves when modifying the value on the first column by the value on the 
corresponding top row of the table. Values above 50\% are highlighted with a
yellow background. For instance, as shown by Table~\ref{table:l}, 94.9\% of the
times that the learning rate parameter goes from 0.05 to 0.025, while keeping 
the rest of the parameters fixed, the average rank across all the analyzed
datasets improves. 

From these tables, we can see what the most favorable parameter values in
general are. In Table~\ref{table:l}, it can be observed that the best values for the
learning rate are intermediate values. Both $0.05$ and $0.1$ clearly improve the
performance of XGBoost with respect to the rest of the parameters on average. 
The best values for {\tt gamma} (see Table~\ref{table:g}) are also in the mid-range
of the analyzed values. In this table, we can observe that the default gamma
value, which is 0, is definitely not the best choice in general. A value of {\tt gamma} 
$\in [0.2,0.3]$ seems to be a reasonable choice. An interesting aspect related to the 
tree depth values, as shown in Table~\ref{table:d}, is that the higher the
depth, the better the performance on average. This is not necessarily in
contradiction with the common use of shallow trees in gradient boosting algorithms
since the depth parameter value is simply a maximum. Furthermore, the actual
depth of the trees is also selected through the gamma parameter (that controls
the complexity of the trees). Regarding the percentage of selected features when
building the tree, it can be observed in Table~\ref{table:f} that values 0.25,
sqrt and log2 perform very similarly on average. As for subsampling,  the best
value is 0.75 (see Table~\ref{table:ss}). In summary, we propose
to use as the default XGBoost parameters: 0.05, 0.2, 100 (unlimited), sqrt and
0.75 for learning rate, gamma, depth, features and subsampling rate
respectively.

\begin{table}
\centering
\begin{tabular}{|c|c|c|c|c|c|}
\hline
\% 	& 0.025 	& 0.05 	& 0.1 	& 0.2 	& 0.3\\
\hline
0.025 	& 0.0 	& 5.1 	& 15.9 	& \cellcolor{yellow!47}52.5 	& \cellcolor{yellow!47}69.0\\
\hline
0.05 	& \cellcolor{yellow!47}94.9 	& 0.0 	& \cellcolor{yellow!47}52.5 	& \cellcolor{yellow!47}77.0 	& \cellcolor{yellow!47}86.5\\
\hline
0.1 	& \cellcolor{yellow!47}84.1 	& 47.5 	& 0.0 	& \cellcolor{yellow!47}85.2 	& \cellcolor{yellow!47}95.2\\
\hline
0.2 	& 47.5 	& 22.9 	& 14.8 	& 0.0 	& \cellcolor{yellow!47}87.0\\
\hline
0.3 	& 31.0 	& 13.5 	& 4.8 	& 13.0 	& 0.0\\
\hline
\end{tabular}
\caption{Percentage of times the average rank of XGBoost improves when changing
the {\tt learning\_rate} from the value in the first column to the value in the 
first row}
\label{table:l}
\end{table}

\begin{table}
\centering
\begin{tabular}{|c|c|c|c|c|c|c|c|c|}
\hline
\% 	& 0.0 	& 0.1 	& 0.2 	& 0.3 	& 0.4 	& 1.0 	& 1.5 	& 2.0\\
\hline
0 	& 0.0 	& \cellcolor{yellow!47}51.7 	& 40.0 	& 43.1 	& 45.4 	& \cellcolor{yellow!47}69.2 	& \cellcolor{yellow!47}85.2 	& \cellcolor{yellow!47}93.8\\
\hline
0.1 	& 48.3 	& 0.0 	& 37.5 	& 43.1 	& 45.0 	& \cellcolor{yellow!47}66.9 	& \cellcolor{yellow!47}83.8 	& \cellcolor{yellow!47}93.3\\
\hline
0.2 	& \cellcolor{yellow!47}60.0 	& \cellcolor{yellow!47}62.5 	& 0.0 	& \cellcolor{yellow!47}51.0 	& \cellcolor{yellow!47}57.9 	& \cellcolor{yellow!47}76.3 	& \cellcolor{yellow!47}90.2 	& \cellcolor{yellow!47}96.7\\
\hline
0.3 	& \cellcolor{yellow!47}56.9 	& \cellcolor{yellow!47}56.9 	& 49.0 	& 0.0 	& 49.6 	& \cellcolor{yellow!47}73.5 	& \cellcolor{yellow!47}91.5 	& \cellcolor{yellow!47}96.7\\
\hline
0.4 	& \cellcolor{yellow!47}54.6 	& \cellcolor{yellow!47}55.0 	& 42.1 	& \cellcolor{yellow!47}50.4 	& 0.0 	& \cellcolor{yellow!47}74.8 	& \cellcolor{yellow!47}91.5 	& \cellcolor{yellow!47}95.8\\
\hline
1 	& 30.8 	& 33.1 	& 23.8 	& 26.5 	& 25.2 	& 0.0 	& \cellcolor{yellow!47}80.8 	& \cellcolor{yellow!47}94.2\\
\hline
1.5 	& 14.8 	& 16.3 	& 9.8 	& 8.5 	& 8.5 	& 19.2 	& 0.0 	& \cellcolor{yellow!47}82.7\\
\hline
2 	& 6.3 	& 6.7 	& 3.3 	& 3.3 	& 4.2 	& 5.8 	& 17.3 	& 0.0\\
\hline
\end{tabular}
\caption{Percentage of times the average rank of XGBoost improves when changing
the {\tt gamma} from the value in the first column to the value in the first row} 
\caption{} \label{table:g}
\end{table}

\begin{table}
\centering
\begin{tabular}{|c|c|c|c|c|c|c|}
\hline
\% 	& 2 	& 3 	& 5 	& 7 	& 10 	& 100\\
\hline
2 	& 0.0 	& 12.8 	& 9.8 	& 7.2 	& 8.1 	& 8.9\\
\hline
3 	& \cellcolor{yellow!47}87.2 	& 0.0 	& 28.1 	& 25.9 	& 23.9 	& 25.0\\
\hline
5 	& \cellcolor{yellow!47}90.2 	& \cellcolor{yellow!47}71.9 	& 0.0 	& 38.1 	& 34.2 	& 33.3\\
\hline
7 	& \cellcolor{yellow!47}92.8 	& \cellcolor{yellow!47}74.1 	& \cellcolor{yellow!47}61.9 	& 0.0 	& 45.6 	& 44.2\\
\hline
10 	& \cellcolor{yellow!47}91.9 	& \cellcolor{yellow!47}76.1 	& \cellcolor{yellow!47}65.8 	& \cellcolor{yellow!47}54.4 	& 0.0 	& 47.0\\
\hline
100 	& \cellcolor{yellow!47}91.1 	& \cellcolor{yellow!47}75.0 	& \cellcolor{yellow!47}66.7 	& \cellcolor{yellow!47}55.8 	& \cellcolor{yellow!47}53.0 	& 0.0\\
\hline
\end{tabular}
\caption{Percentage of times the average rank of XGBoost improves when changing
the {\tt depth} from the value in the first column to the value in the first row} 
\label{table:d}
\end{table}

\begin{table}
\centering
\begin{tabular}{|c|c|c|c|c|}
\hline
\% 	& 0.25 	& sqrt 	& log2 	& 1.0\\
\hline
0.25 	& 0.0 	& 44.4 	& 44.9 	& \cellcolor{yellow!47}67.2\\
\hline
sqrt 	& \cellcolor{yellow!47}55.6 	& 0.0 	& \cellcolor{yellow!47}52.0 	& \cellcolor{yellow!47}72.3\\
\hline
log2 	& \cellcolor{yellow!47}55.1 	& 48.0 	& 0.0 	& \cellcolor{yellow!47}70.9\\
\hline
1 	& 32.8 	& 27.7 	& 29.1 	& 0.0\\
\hline
\end{tabular}
\caption{Percentage of times the average rank of XGBoost improves when changing
the {\tt colsample\_bylevel} from the value in the first column to the value in the first row} 
\label{table:f}
\end{table}

\begin{table}
\centering
\begin{tabular}{|c|c|c|c|c|}
\hline
\% 	& 0.25 	& 0.5 	& 0.75 	& 1.0\\
\hline
0.25 	& 0.0 	& 5.3 	& 7.8 	& 22.4\\
\hline
0.5 	& \cellcolor{yellow!47}94.7 	& 0.0 	& 30.5 	& 47.7\\
\hline
0.75 	& \cellcolor{yellow!47}92.2 	& \cellcolor{yellow!47}69.5 	& 0.0 	& \cellcolor{yellow!47}75.4\\
\hline
1 	& \cellcolor{yellow!47}77.6 	& \cellcolor{yellow!47}52.3 	& 24.6 	& 0.0\\
\hline
\end{tabular}
\caption{Percentage of times the average rank of XGBoost improves when changing
the {\tt subsample} from the value in the first column to the value in the first row} 
\label{table:ss}
\end{table}

In Table~\ref{table:accuracy2}, the average error for the proposed parameters is
shown in column ``Prop. Def.". For reference, the default parametrization is also
shown in the first column. 
As it can be observed, the proposed fixed parametrization
improves over the results of the default setting. With the proposed  default setting,
better results can be obtained in 17 out of 28 datasets with notable differences in datasets as 
{\it Echo}, {\it Sonar} or {\it Tic-tac-toe}. In addition, when the default
setting is better than the proposed parameters, the differences are in general small,
except for {\it German}, {\it Parkinson} and {\it Vehicle}.

\begin{figure}
\centering
\rotatebox{270}{\includegraphics[scale=0.7]{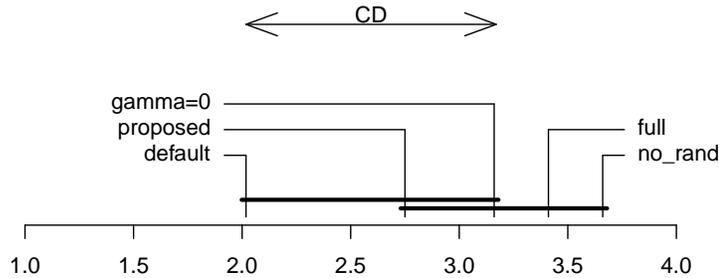}}
\caption{Average ranks (higher rank is better) for different XGBoost
configurations (Critical difference CD= 1.15)}
\label{demsar2}
\end{figure}

In spite of the improvements on average achieved by the proposed parameter
setting, it seems clear that parameter optimization is necessary to further
improve the performance of XGBoost and to adapt the model to the characteristics
of each specific dataset. 
In this context, we carry out an experiment to explore two different parameter grids. On
one hand, we would like to analyze the differences between gradient boosting and
XGBoost in more detail. There are little differences between both algorithms except that
XGBoost is optimized for speed. The main difference, from a machine learning
point of view, is that
XGBoost incorporates into the loss function a parameter to explicitly control
the complexity of the decision trees (i.e. gamma). In order to analyze whether
this parameter provides any advantage in the
classification performance of XGBoost, a grid with the same parameter values as the ones
given by Table~\ref{table:params} is used except for gamma, which is always set 
to $0.0$. On the other hand, we have observed that, in the case of random forest,
tuning the randomization parameters is not very productive in general. As shown
in Figure~\ref{demsar}, the average rank of default random forest is
better than the rank of the forests for which the randomization parameters are
tuned. Hence, the second proposed grid is to tune the optimization
parameters of XGBoost (i.e. learning rate,
gamma and depth) keeping the randomization parameters (i.e. random
features and subsampling) fixed. The
randomization parameters will be fixed to 0.75 for the subsampling ratio and
to sqrt for the number of features as suggested by Tables~\ref{table:ss} and
\ref{table:f} respectively. The average generalization errors for XGBoost when using these 
two grids are shown in Table~\ref{table:accuracy2}. Column ``No gamma" shows the
results for the grid that does not tune gamma (i.e. gamma=0), and column 
``No rand tun." shows the results for the grid that does not tune the randomization
parameters.
The best average test error for each dataset is highlighted with a different
background color. Additionally, the average ranks for this table are
shown in Figure~\ref{demsar2} following the methodology proposed in \cite{demsar2006}. 
In this figure, differences among methods connected with a horizontal solid
line are not statistically significant.

\begin{sidewaystable}
\centering
\begin{tabular}{| c | c | c | c | c | c |}
\hline
\textbf{Dataset} 	& \textbf{Default} 	& \textbf{Prop. Def.} 	& \textbf{Tuned} 	&
\textbf{No rand tun.} 	& \textbf{No gamma} \\
\hline
Australian	& 86.94\%$\pm$2.73	& \cellcolor{yellow!47}88.11\%$\pm$3.08	& 87.53\%$\pm$3.47	& 87.54\%$\pm$3.00	& 86.95\%$\pm$3.62\\
\hline
Banknote	& 99.64\%$\pm$0.59	& 99.42\%$\pm$0.54	& 99.64\%$\pm$0.49	& 99.56\%$\pm$0.58	& \cellcolor{yellow!47}99.71\%$\pm$0.48\\
\hline
Breast	& 96.28\%$\pm$1.15	& 96.28\%$\pm$1.15	& 95.99\%$\pm$1.68	& \cellcolor{yellow!47}96.99\%$\pm$1.50	& 95.99\%$\pm$1.55\\
\hline
Cleveland	& 81.14\%$\pm$7.25	& 82.78\%$\pm$6.24	& \cellcolor{yellow!47}83.16\%$\pm$7.94	& 82.12\%$\pm$6.69	& 83.09\%$\pm$8.53\\
\hline
Dermatology	& 96.74\%$\pm$3.21	& 97.27\%$\pm$3.24	& 97.27\%$\pm$3.24	& \cellcolor{yellow!47}98.39\%$\pm$3.50	& 98.08\%$\pm$3.29\\
\hline
Diabetes	& 75.65\%$\pm$5.11	& 75.91\%$\pm$4.30	& 76.56\%$\pm$4.50	& 76.17\%$\pm$4.43	& \cellcolor{yellow!47}77.08\%$\pm$4.51\\
\hline
Echo	& 94.46\%$\pm$6.80	& \cellcolor{yellow!47}98.75\%$\pm$3.75	& \cellcolor{yellow!47}98.75\%$\pm$3.75	& \cellcolor{yellow!47}98.75\%$\pm$3.75	& \cellcolor{yellow!47}98.75\%$\pm$3.75\\
\hline
Ecoli	& 86.87\%$\pm$5.31	& 87.17\%$\pm$6.50	& \cellcolor{yellow!47}89.05\%$\pm$4.12	& 87.50\%$\pm$5.88	& 87.81\%$\pm$5.48\\
\hline
German	& \cellcolor{yellow!47}79.00\%$\pm$4.22	& 75.90\%$\pm$4.53	& 77.40\%$\pm$4.13	& 77.30\%$\pm$4.22	& 77.20\%$\pm$3.28\\
\hline
Heart	& 79.26\%$\pm$5.29	& 81.85\%$\pm$5.84	& 84.07\%$\pm$5.98	& 84.44\%$\pm$6.15	& \cellcolor{yellow!47}84.81\%$\pm$4.81\\
\hline
Hepatitis	& 59.21\%$\pm$8.28	& 59.83\%$\pm$9.75	& 67.00\%$\pm$6.56	& \cellcolor{yellow!47}67.50\%$\pm$11.82	& 66.92\%$\pm$11.04\\
\hline
Ionosphere	& 92.56\%$\pm$2.69	& \cellcolor{yellow!47}93.72\%$\pm$2.51	& 92.59\%$\pm$3.17	& 92.87\%$\pm$3.43	& 91.99\%$\pm$3.37\\
\hline
Iris	& 92.67\%$\pm$6.29	& \cellcolor{yellow!47}95.33\%$\pm$4.27	& 94.00\%$\pm$4.67	& \cellcolor{yellow!47}95.33\%$\pm$4.27	& 94.00\%$\pm$5.54\\
\hline
Liver	& 68.65\%$\pm$4.69	& 68.98\%$\pm$5.90	& 68.11\%$\pm$6.12	& \cellcolor{yellow!47}69.97\%$\pm$6.13	& 69.31\%$\pm$5.60\\
\hline
Magic04	& 87.47\%$\pm$0.57	& \cellcolor{yellow!47}88.66\%$\pm$0.45	& 88.63\%$\pm$0.48	& 88.62\%$\pm$0.31	& 88.46\%$\pm$0.47\\
\hline
New-thyroid	& \cellcolor{yellow!47}95.80\%$\pm$3.26	& 95.35\%$\pm$3.58	& \cellcolor{yellow!47}95.80\%$\pm$3.26	& 95.30\%$\pm$3.01	& 94.85\%$\pm$3.29\\
\hline
Parkinsons	& \cellcolor{yellow!47}92.14\%$\pm$5.72	& 91.70\%$\pm$4.35	& 90.14\%$\pm$4.01	& 90.70\%$\pm$3.37	& 90.14\%$\pm$4.72\\
\hline
Phishing	& 89.65\%$\pm$2.44	& 89.07\%$\pm$2.30	& \cellcolor{yellow!47}91.13\%$\pm$1.30	& 90.69\%$\pm$1.99	& 90.03\%$\pm$2.22\\
\hline
Segment	& 98.48\%$\pm$0.70	& 98.40\%$\pm$0.87	& \cellcolor{yellow!47}98.70\%$\pm$0.77	& 98.66\%$\pm$0.81	& 98.66\%$\pm$0.79\\
\hline
Sonar	& 85.59\%$\pm$6.44	& \cellcolor{yellow!47}87.52\%$\pm$8.17	& 86.97\%$\pm$4.84	& 87.50\%$\pm$6.29	& 87.02\%$\pm$7.45\\
\hline
Soybean	& 94.80\%$\pm$3.35	& 94.65\%$\pm$3.28	& 95.22\%$\pm$2.96	& 94.65\%$\pm$3.05	& \cellcolor{yellow!47}95.53\%$\pm$3.12\\
\hline
Spambase	& 95.17\%$\pm$1.29	& 95.67\%$\pm$1.06	& 95.57\%$\pm$1.28	& 95.65\%$\pm$1.20	& \cellcolor{yellow!47}95.76\%$\pm$1.27\\
\hline
Teaching	& 63.55\%$\pm$8.00	& 64.21\%$\pm$9.54	& 64.26\%$\pm$14.15	& \cellcolor{yellow!47}66.84\%$\pm$10.86	& 63.55\%$\pm$11.86\\
\hline
Tic-tac-toe	& 96.56\%$\pm$1.63	& 99.58\%$\pm$0.51	& \cellcolor{yellow!47}100.00\%$\pm$0.00	& \cellcolor{yellow!47}100.00\%$\pm$0.00	& \cellcolor{yellow!47}100.00\%$\pm$0.00\\
\hline
Vehicle	& 78.74\%$\pm$3.29	& 76.95\%$\pm$2.69	& 77.18\%$\pm$2.31	& \cellcolor{yellow!47}79.21\%$\pm$2.80	& 77.79\%$\pm$2.83\\
\hline
Vowel	& 92.63\%$\pm$3.53	& 94.34\%$\pm$2.76	& \cellcolor{yellow!47}95.86\%$\pm$2.14	& 94.95\%$\pm$2.12	& 95.56\%$\pm$2.31\\
\hline
Waveform	& 85.72\%$\pm$1.05	& 85.54\%$\pm$1.37	& 85.72\%$\pm$1.77	& \cellcolor{yellow!47}85.78\%$\pm$1.23	& 85.56\%$\pm$1.80\\
\hline
Wine	& 97.18\%$\pm$2.83	& \cellcolor{yellow!47}98.82\%$\pm$2.37	& \cellcolor{yellow!47}98.82\%$\pm$2.37	& \cellcolor{yellow!47}98.82\%$\pm$2.37	& \cellcolor{yellow!47}98.82\%$\pm$2.37\\
\hline
\end{tabular}
\caption{Average accuracy and standard deviation of XGBoost with different
configurations: default, proposed, tuned, no gamma tuning and no randomizations
parameter tuning
} \label{table:accuracy2}
\end{sidewaystable}

The results shown in Table~\ref{table:accuracy2} and Figure~\ref{demsar2} are
quite interesting. The number of best results are 11, 9 and 8 when the grid
without randomization parameter tuning, the full grid and the grid without gamma 
tuning are applied respectively. Similarly, the average ranks for these three
methods, shown in Figure~\ref{demsar2}, is favorable to the grid without
randomization parameter tuning, then for the full grid and finally the grid without gamma tuning.
One tendency that is observed from these results is that it seems that including
a complexity term to control the size of the trees can have a small edge over
not using it, although the differences are not statistically significant. A
conclusion that may be clearer is that it seems unnecessary to tune the number
of random features and the subsampling rate provided that those techniques are
applied with reasonable values (in our case subsampling to 0.75 and feature
sampling to sqrt).

Finally, the time required to perform the grid search and to train
the single models is shown in Table~\ref{table:time2} in the same manner as
Table~\ref{table:time}. As shown in the last row of this table for the tested
settings, performing the grid search without tuning the randomization parameters
is over 16 times faster the tuning the full grid on average. These results 
reinforce the fact that tuning the randomization parameter is unnecessary. 
 
\begin{table}
\centering
\begin{adjustbox}{width=1\textwidth}
\begin{tabular}{| c | c | c | c | c | c |}
\hline
\textbf{Dataset} 	& \textbf{Default} 	& \textbf{Proposed} 	& \textbf{Tuned} 	&
\textbf{No gamma} 	& \textbf{No rand tun.} \\
\hline
Australian 	& 0.10 	& 0.09 	& 2775 + 0.08 	& 149 + 0.08 	& 340 + 0.11\\
\hline
Banknote 	& 0.10 	& 0.11 	& 2874 + 0.06 	& 377 + 0.06 	& 340 + 0.06\\
\hline
Breast 	& 0.06 	& 0.06 	& 1733 + 0.04 	& 108 + 0.05 	& 212 + 0.04\\
\hline
Cleveland 	& 0.05 	& 0.04 	& 1357 + 0.02 	& 78 + 0.03 	& 165 + 0.02\\
\hline
Dermatology 	& 0.46 	& 0.28 	& 11014 + 0.27 	& 572 + 0.27 	& 1281 + 0.27\\
\hline
Diabetes 	& 0.08 	& 0.11 	& 2982 + 0.06 	& 163 + 0.08 	& 365 + 0.05\\
\hline
Echo 	& 0.02 	& 0.01 	& 532 + 0.01 	& 33 + 0.01 	& 65 + 0.01\\
\hline
Ecoli 	& 0.17 	& 0.15 	& 4518 + 0.11 	& 284 + 0.13 	& 538 + 0.11\\
\hline
German 	& 0.17 	& 0.19 	& 5453 + 0.16 	& 251 + 0.13 	& 670 + 0.19\\
\hline
Heart 	& 0.05 	& 0.05 	& 1268 + 0.03 	& 74 + 0.03 	& 155 + 0.02\\
\hline
Hepatitis 	& 0.04 	& 0.03 	& 1071 + 0.02 	& 61 + 0.02 	& 131 + 0.02\\
\hline
Ionosphere 	& 0.14 	& 0.06 	& 2487 + 0.05 	& 112 + 0.04 	& 289 + 0.04\\
\hline
Iris 	& 0.04 	& 0.04 	& 1308 + 0.03 	& 171 + 0.04 	& 154 + 0.03\\
\hline
Liver 	& 0.07 	& 0.09 	& 2308 + 0.05 	& 129 + 0.05 	& 283 + 0.06\\
\hline
Magic04 	& 3.32 	& 8.07 	& 123764 + 7.86 	& 6618 + 7.38 	& 14974 + 8.11\\
\hline
New-thyroid 	& 0.05 	& 0.05 	& 1728 + 0.04 	& 112 + 0.05 	& 198 + 0.04\\
\hline
Parkinsons 	& 0.05 	& 0.03 	& 1169 + 0.03 	& 64 + 0.02 	& 138 + 0.03\\
\hline
Phishing 	& 0.36 	& 0.47 	& 12464 + 0.75 	& 746 + 0.40 	& 1551 + 0.53\\
\hline
Segment 	& 3.22 	& 2.13 	& 73309 + 2.21 	& 3878 + 1.48 	& 8447 + 2.25\\
\hline
Sonar 	& 0.15 	& 0.05 	& 2442 + 0.05 	& 103 + 0.04 	& 280 + 0.06\\
\hline
Soybean 	& 3.73 	& 2.80 	& 104873 + 2.69 	& 5892 + 2.65 	& 12730 + 3.49\\
\hline
Spambase 	& 1.83 	& 1.44 	& 47412 + 1.46 	& 1621 + 1.20 	& 5788 + 1.54\\
\hline
Teaching 	& 0.05 	& 0.07 	& 1879 + 0.06 	& 124 + 0.06 	& 228 + 0.05\\
\hline
Tic-tac-toe 	& 0.08 	& 0.10 	& 2842 + 0.08 	& 171 + 0.08 	& 347 + 0.09\\
\hline
Vehicle 	& 0.60 	& 0.59 	& 17217 + 0.38 	& 899 + 0.28 	& 1993 + 0.35\\
\hline
Vowel 	& 1.96 	& 1.60 	& 48726 + 1.24 	& 2887 + 1.40 	& 5577 + 1.24\\
\hline
Waveform 	& 3.34 	& 4.13 	& 111562 + 1.74 	& 5028 + 1.73 	& 13073 + 2.15\\
\hline
Wine 	& 0.07 	& 0.06 	& 2104 + 0.05 	& 122 + 0.05 	& 231 + 0.05\\
\hline
\hline
Ave. ratio 	&1.0 	& 1.0 	& 29043.7 + 0.8 	& 1782.3 + 0.8 	& 3472.0 + 0.8 \\ 
\hline
\end{tabular}
\end{adjustbox}
\caption{Average execution time (in seconds) for training XGBoost,
random forest and gradient boosting (more details in the text)}
\label{table:time2}
\end{table}

\section{Conclusion}
In this study we present an empirical analysis of XGBoost, a method based on
gradient boosting that has proven to be an efficient challenge solver. 
Specifically, the performance of XGBoost in terms of training speed and
accuracy is compared with the performance of gradient boosting and random
forest under a wide range of classification tasks. In addition, the parameter
tuning process of XGBoost is thoroughly analyzed.

The results of this study show that the most accurate classifier, in terms of
the number of problems with the best performance in the problems investigated, was
gradient boosting. Nevertheless, the differences with respect to XGBoost and to
random forest using the default parameters are not statistically significant in
terms of average ranks. We observed that XGBoost and gradient boosting trained
using the default parameters of the packages were the least successful methods. 
In consequence, we conclude that a meticulous parameter search is necessary to
create accurate models based on gradient boosting. This is not the case for
random forest, whose generalization performance was slightly better on average
when the default parameter values were used (those originally proposed by Breiman).
In fact, tuning in XGBoost the randomization parameters subsampling rate and
the number of features selected at each split was found to be unnecessary as long as
some randomization is used. In our experiments, we fixed the values of
the subsampling rate to 0.75 without replacement and the number of features to sqrt,
reducing the size of the parameter grid search 16 fold and improving the average
performance of XGBoost.

Finally, from the experiments of this
study, which are based on grid search parameter tuning using within-train
10-fold cross-validation, the tuning phase contributed to over 99.9\% of the
computational effort necessary to train gradient boosting or XGBoost. The grid search time can however be dramatically reduced when the smaller proposed grid
is used for XGBoost.

These results are not necessarily in contradiction with the top performances
obtained by XGBoost in Kaggle competitions. The best contender in such
competitions is the single model that achieves the best performance even if it is
only for a slight margin. XGBoost allows for a fine parameter tuning using a
computationally efficient algorithm. This is not as feasible with random forest
(as small gains are obtained, if at all, with parameter tuning) or with gradient
boosting, which requires longer computational times. 

\section*{Acknowledgements}
The authors acknowledge financial support from
the European Regional Development Fund and from the
Spanish Ministry of Economy, Industry, and
Competitiveness - State Research
Agency, project TIN2016-76406-P (AEI/FEDER, UE)

\bibliographystyle{plain}
\bibliography{refs}

\end{document}